\documentclass[10pt,twocolumn]{article} 
\usepackage{simpleConference}
\usepackage{times}
\usepackage{graphicx}
\usepackage{amssymb}
\usepackage{amsthm}
\usepackage{url,hyperref}
\usepackage{hyperref}
\usepackage{url}
\usepackage[ruled,noend]{algorithm2e}
\usepackage{amsmath}
\usepackage{amsfonts}
\usepackage{graphicx}
\usepackage{comment}
\usepackage{subcaption}

\usepackage{booktabs}
\usepackage{array}
\usepackage{multirow}
\usepackage{enumitem}
\usepackage{makecell}
\usepackage{pifont}
\usepackage{wrapfig}

\usepackage{listings}  % For lstlisting
\usepackage{xcolor}    % For syntax highlighting

\lstdefinelanguage{json}{
    string=[s]{"}{"},
    comment=[l]{//},
    morecomment=[s]{/*}{*/},
}

\usepackage{times}  % DO NOT CHANGE THIS
\usepackage{helvet}  % DO NOT CHANGE THIS
\usepackage{courier}  % DO NOT CHANGE THIS
\usepackage{graphicx} % DO NOT CHANGE THIS
\usepackage{natbib}  % DO NOT CHANGE THIS AND DO NOT ADD ANY OPTIONS TO IT
\usepackage{caption} % DO NOT CHANGE THIS AND DO N
\usepackage{xcolor}
\usepackage{lipsum}

\newcommand{\ourmodel}{\textsc{NEMO-4-PAYPAL}\xspace}

\usepackage{xcolor}

\begin{document}

\title{\ourmodel: Leveraging NVIDIA’s Nemo Framework for empowering PayPal’s Commerce Agent  }

% \author{
% PayPal AI and Nvidia AI  \\
% USA
% }

% \author{
% AAA \\
% AA@gmail.com\\
% SF \\
% USA
% \and 
% BB \\
% bb@gmail.com\\
% CC \\
% USA
% \and 
% SS\\
% mm@uic.edu\\
% UU\\
% USA\\
% \and 
% SC\\
% gg@edu.qa\\
% Computing Research Institute\\
% Qatar
% }

\author{
Sudhanshu Garg$^1$, 
Andrew Wang$^1$, 
Chaitanya Kulkarni$^1$ ,
Ali Sahami$^1$\\[0.5em]
Farhad Farahani$^1$, 
Sean Yun-Shiuan Chuang$^1$, 
Jian Wan$^1$, 
Srinivasan Manoharan$^1$ \\[0.5em]
Uma Kona$^1$,
Nitin Sharma$^1$,
Linsey Pang$^1$,
Prakhar Mehrotra$^1$ \\[0.5em]
Jessica Clark$^2$, 
Mark Moyou$^2$
}
\affiliation{
$^1$PayPal AI\\
$^2$NVIDIA
}

% \author{
% Ali Sahami \\
% qzjiang21@m.fudan.edu.cn\\
% FuDan University \\
% China
% \and 
% Sudhanshu Garg\\
% panglinsey@gmail.com\\
% Salesforce \\
% USA
% \and 
% Andrew Wang \\
% alicegatti1992@gmail.com\\
% Center for AI Safety \\
% USA
% \and 
% Chaitanya Kulkarni \\
% maggarwal@hbku.edu.qa\\
% Qatar Computing Research Institute, HBKU \\
% Qatar
% \and 
% Farhad Farahani\\
% gvantini@hbku.edu.qa\\
% Qatar Computing Research Institute, HBKU \\
% Qatar
% \and 
% Sean Yun-Shiuan Chuang \\
% gvantini@hbku.edu.qa\\
% Qatar Computing Research Institute, HBKU \\
% Qatar
% \and 
% Jian Wan\\
% Xiaosong.Ma@mbzuai.ac.ae\\
% Mohamed bin Zayed University \\
% United Arab Emirates
% \and 
% Srinivasan Manoharan\\
% wwsun@fudan.edu.cn\\
% FuDan University \\
% China
% \and 
% Uma K\\
% wwsun@fudan.edu.cn\\
% FuDan University \\
% China
% Linsey Pang\\
% wwsun@fudan.edu.cn\\
% FuDan University \\
% China
% \and 
% Prakhar Mehrotra\\
% wwsun@fudan.edu.cn\\
% FuDan University \\
% China
% \and 
% Jessica Clark\\
% schawla@hbku.edu.qa\\
% Qatar Computing Research Institute, HBKU \\
% Qatar
% \and 
% Mark Moyou\\
% wwsun@fudan.edu.cn\\
% FuDan University \\
% China
% }

\maketitle

\begin{abstract}

We present the development and optimization of PayPal's Commerce Agent, powered by \ourmodel, a multi-agent system designed to revolutionize agentic commerce on the PayPal platform. Through our strategic partnership with NVIDIA, we leveraged the NeMo Framework for LLM model fine-tuning to enhance agent performance. Specifically, we optimized the Search and Discovery agent by replacing our base model with a fine-tuned Nemotron small language model (SLM).
 We conducted comprehensive experiments using the llama3.1-nemotron-nano-8B-v1 architecture, training LoRA-based models through systematic hyperparameter sweeps across learning rates, optimizers (Adam, AdamW), cosine annealing schedules, and LoRA ranks. Our contributions include: (1) the first application of NVIDIA's NeMo Framework to commerce-specific agent optimization, (2) LLM powered fine-tuning strategy for retrieval-focused commerce tasks, (3) demonstration of significant improvements in latency and cost while maintaining agent quality, and (4) a scalable framework for multi-agent system optimization in production e-commerce environments. Our results demonstrate that the fine-tuned Nemotron SLM effectively resolves the key performance issue in the retrieval component, which represents over 50\% of total agent response time, while maintaining or enhancing overall system performance.

\end{abstract}
\begin{figure*}[!t]
\centering
\vspace{-1mm}
\includegraphics[width=0.85\textwidth]{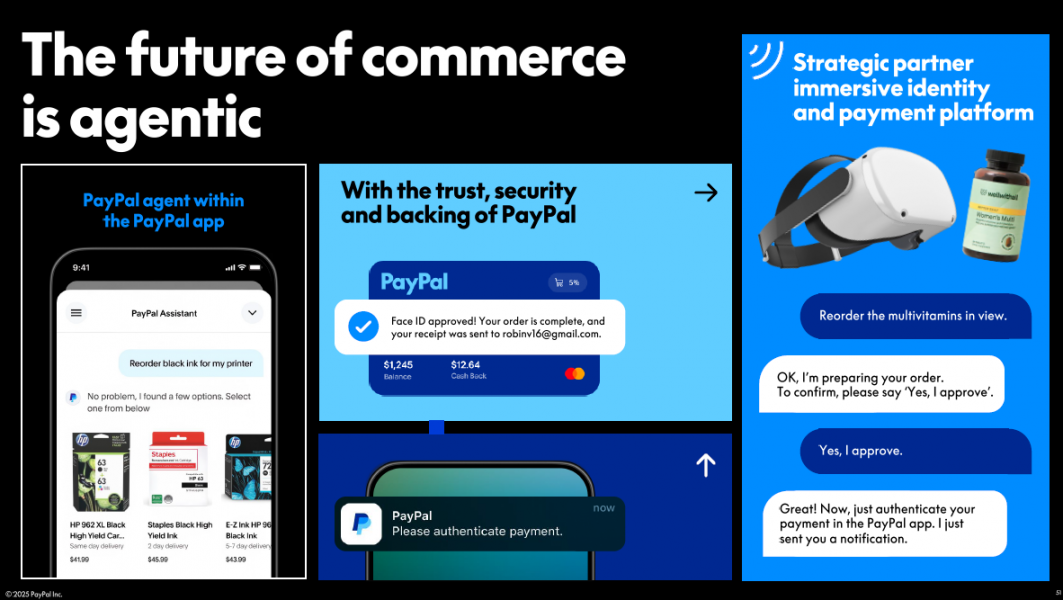}
\vspace{-1.5mm}
\caption{PayPal is transforming from a payment company to a commerce platform with, our commerce Paypal Agent, which leverages NVIDIA’s Nemo. }
\vspace{-2mm}
\label{fig:teaser}
\end{figure*}

\section{Introduction}
% \label{sec:Introduction}

The emergence of agentic artificial intelligence represents a paradigm shift in how consumers interact with e-commerce platforms \cite{purcarea2025commerce, kodali2025agentic}, promising to transform the traditional shopping experience into an intelligent, conversational journey. PayPal's Commerce Agent embodies this vision by enabling autonomous shopping assistance across the entire commerce lifecycle, from pre-purchase discovery through post-purchase support. This multi-agent system leverages advanced language models to provide personalized shopping experiences, positioning itself at the forefront of the agentic commerce revolution.

The complexity of modern e-commerce interactions demands sophisticated AI systems capable of understanding nuanced customer intent, navigating vast product catalogs, and orchestrating seamless transactions. Traditional rule-based systems and simple chatbots fail to meet these requirements, necessitating the development of intelligent agents powered by large language models (LLMs) \cite{goldrevolutionizing}. However, the deployment of such systems in production environments presents significant challenges in terms of latency, cost, and performance optimization, particularly when handling the scale and diversity of PayPal's global commerce ecosystem.

% Our initial implementation of the Commerce Agent utilized OpenAI's GPT-4o \cite{openai_gpt4o_2024} as the foundational language model, providing robust conversational capabilities and general-purpose reasoning. However, extensive performance analysis showed critical optimization opportunities, particularly in the retrieval component of our Search and Discovery agent. With retrieval operations accounting for over 50\% of total system latency and serving as a primary determinant of agent quality, this component became our primary focus for optimization efforts.

% We present our strategic collaboration with NVIDIA to address these performance challenges through the application of the NeMo Framework \cite{NVIDIA_NeMo_Framework}, a state-of-the-art platform for training and deploying specialized language models. Our approach centers on replacing general-purpose LLMs with domain-specific, fine-tuned small language models (SLMs) optimized for commerce retrieval tasks. Specifically, we leverage NVIDIA's Nemotron model family, enterprise-ready architectures designed for high-performance inference and specialized fine-tuning.

Our first version of the Commerce Agent provided good conversation abilities and general reasoning skills. However, detailed performance testing revealed important areas for improvement, especially in the search component of our Search and Discovery agent. Since search operations took over 50\% of total system response time and directly affected agent quality, we focused our optimization work on this component.

We collaborated with NVIDIA to solve these performance challenges by leveraging the NeMo Framework \cite{NVIDIA_NeMo_Framework} and Nemotron model family with specialized, fine-tuned small language models (SLMs) optimized for commerce search tasks.

The contributions of this work are significant for the industrial AI communities: 
\begin{itemize}[itemsep=0pt, parsep=0pt]
\item \textbf{NeMo Framework Application}: We demonstrate the first application of NVIDIA's NeMo Framework to commerce-specific agent optimization, establishing a systematic methodology for model selection and fine-tuning in production e-commerce environments.
\item \textbf{LLM powered Fine-tuning Strategy}: Instead of implementing traditional IR method or prompting pre-trained LLM model for search and recommendation,  we  fine-tuned small language models (SLMs) optimized for commerce retrieval tasks. 

\item \textbf{Performance Optimization Evidence}: We conducted experiments with 20 LoRA-based model variants exploring hyperparameter spaces across learning rates, optimizers, scheduling strategies, and architectural configurations. The fine-tuned model demonstrates significant improvements in latency and operational cost while maintaining or improving overall system performance metrics.

\item \textbf{Scalable Multi-Agent Framework}: We establish a scalable framework for multi-agent system optimization applicable across various components of complex AI systems, providing a blueprint for future enhancements and extensions.
\end{itemize}
The remainder of this paper is organized as follows: Section II reviews related work in agentic systems, traditional IR system and Nvidia's Nemo Framework. Section III details our commerce search and recommendation components and multi-agent architecture. Section V describes our experimental approach, covering dataset preparation, model fine-tuning, and inference testing procedures. We present comprehensive results and analysis from our fine-tuning experiments, as well as comparisons between different optimization SDK frameworks. Finally,  we give  our conclusion and the future work in Section V.

\section{Related Work} %(1 pg, including conclusion)
\label{sec:related-work}

\begin{figure*}[th]
\centering
\vspace{-1mm}
\includegraphics[width=0.85\textwidth]{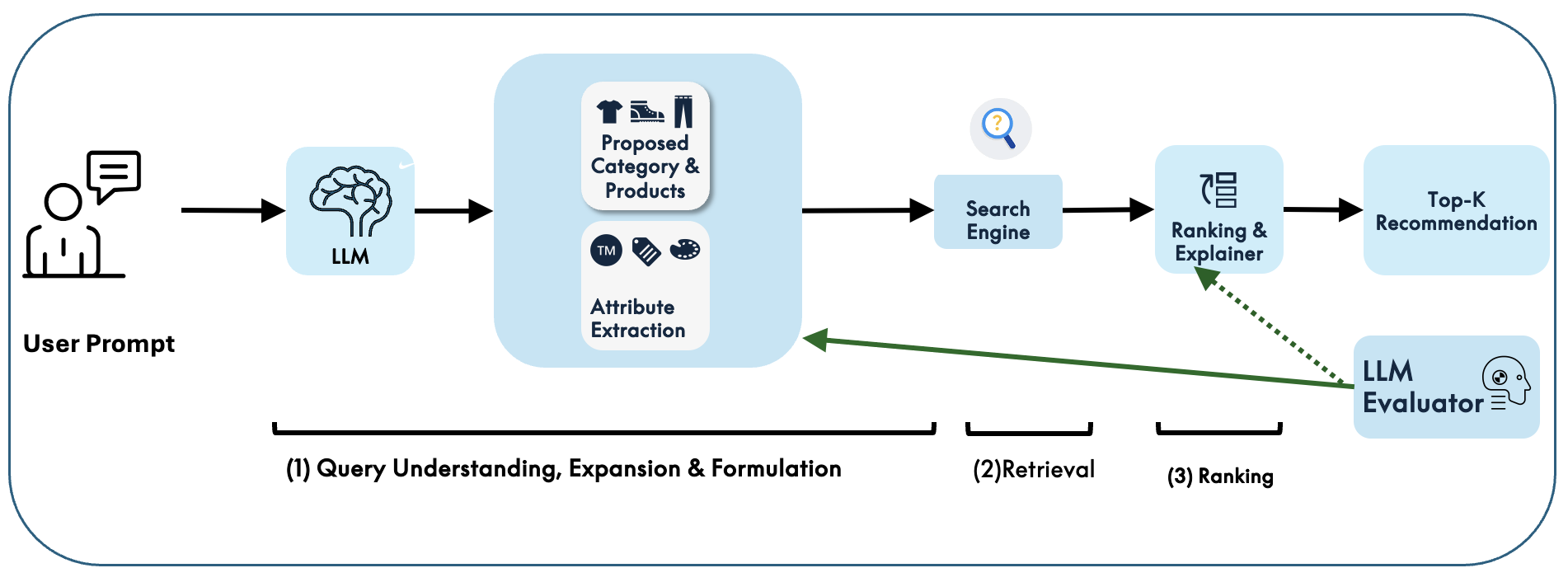} 
% irnow.png}
\vspace{-1.5mm}
\caption{Commerce Search and Recommendation System at PayPal.
The system comprises four key components: (1) Query Understanding, Expansion, and Formulation (LLM); (2) Retrieval; (3) Ranking; and (4) LLM Evaluator. }
\vspace{-2mm}
\label{fig:recsystem}
\end{figure*}

\subsection{E-Commerce Information Retrieval  }
Traditional e-commerce information retrieval systems \cite{manning_introduction_2008} have predominantly relied on structured data representations, operating over well-defined product catalogs, taxonomies, and attribute-based schemas to enable search and recommendation functionalities. The conventional approach involves a systematic pipeline of converting unstructured textual data---such as product descriptions, user reviews, and query inputs---into structured formats through information extraction techniques, followed by conducting search operations over these normalized data structures. This methodology leverages relational databases, inverted indices, and rule-based matching algorithms to achieve precise retrieval based on explicit attribute matching and categorical filtering. While this structured approach provides deterministic results and enables efficient query processing through established indexing mechanisms, it faces inherent limitations in handling semantic nuances, contextual understanding, and the growing complexity of natural language queries that characterize modern e-commerce interactions \cite{wang2024rethinking}. 

Recent advances in large language models (LLMs) have demonstrated strong capabilities in natural language understanding and generation, showing new opportunities for bridging the gap between unstructured user queries and structured product representations \cite{gao2023precise}. These models can effectively extract user intent and product attributes from conversational queries, and reformulate ambiguous or underspecified searches into precise, structured formats that align with product catalog schemas, thereby enhancing retrieval effectiveness in e-commerce systems \cite{zhang2025leser,jagerman2023query}.

\subsection{LLM Agent in Search and Recommendation}
LLM agents \cite{huang2024understanding} are transforming commerce search and recommendation systems by enabling more natural, context-aware interactions between users and e-commerce platforms. Unlike traditional keyword-based search systems that rely on exact matches and predefined filters, LLM agents leverage large language models to understand user intent across multiple dimensions, including implicit preferences, contextual constraints, and evolving needs throughout a shopping session. These agents perform sophisticated query understanding by extracting attributes (e.g., brand, color, price range, size), expanding queries with related terms and synonyms, and reformulating ambiguous requests into structured search parameters. Beyond search, LLM agents orchestrate multi-step recommendation workflows by reasoning over product catalogs, user purchase history, and real-time session data to generate personalized suggestions. They handle complex conversational scenarios such as comparison requests ("show me Nike shoes similar to what I bought last month but in blue"), constraint satisfaction ("find running shoes under $\$100$ with good arch support"), and cross-category exploration ("I need shoes for hiking, what else should I get for my trip?"). By integrating retrieval systems with ranking models, LLM agents ground their responses in actual product inventory while maintaining conversational fluency, ultimately reducing search friction, increasing discovery, and improving conversion rates in modern commerce platforms.

\subsection{NVIDIA’s NeMo Framework}
NVIDIA NeMo Framework \cite{NVIDIA_NeMo_Framework} is a complete AI platform that helps researchers and developers work with Large Language Models (LLMs), multimodal systems, and speech AI applications. The framework supports the entire process of building AI models, from preparing data and training models to putting them into production, and can run on different computing environments including local servers, data centers, or cloud services. NeMo includes tools for data preparation, efficient training methods like supervised fine-tuning and LoRA (a technique that reduces training time and resources), and multiple options for deploying trained models.

\begin{figure*}[th]
\centering
\vspace{-1mm}
\includegraphics[width=0.85\textwidth]{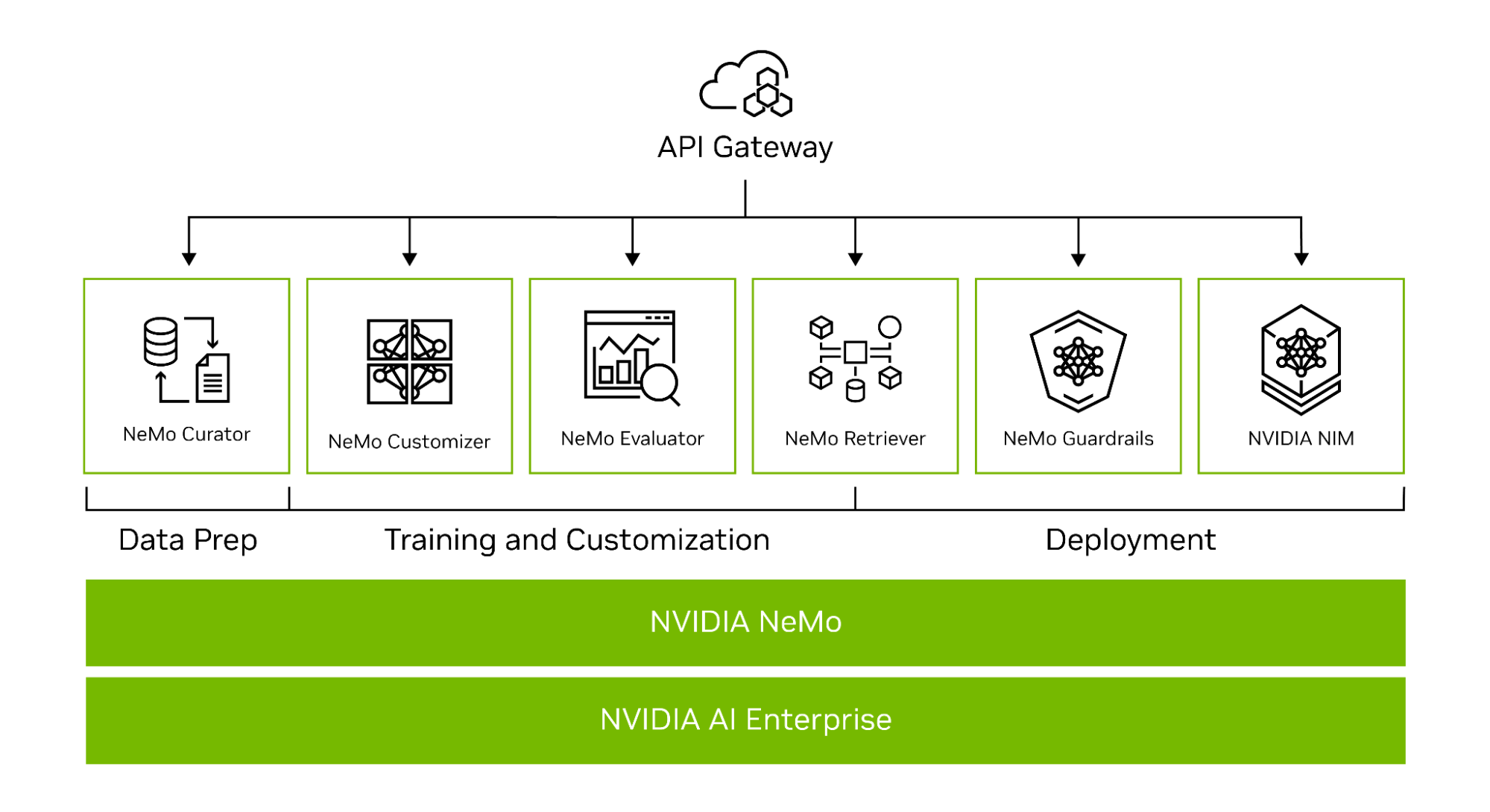}
\vspace{-1.5mm}
\caption{NeMo Framework \cite{nvidia_nemo_framework_2024} NVIDIA NeMo is an end-to-end platform for developing custom generative AI, anywhere. It includes tools for training, fine-tuning, retrieval-augmented generation, guardrailing, data curation, and pretrained models. It has offerings across the tech stack, from frameworks to high-level API endpoints.}
\vspace{-2mm}
\label{fig:nemo}
\end{figure*}

The framework works with many popular model types including LLama, Mistral, GPT, and other major language models, plus vision and speech models for tasks like speech recognition and text-to-speech. Built using PyTorch and Python, NeMo provides ready-to-use model templates, tested training recipes, and infrastructure that can scale across multiple GPUs and machines, making it well-suited for large-scale AI projects and enterprise use. (see Figure \ref{fig:nemo})

\section{\ourmodel Framework}
This section presents the key components of our \ourmodel framework.
\begin{figure*}[t]
\centering
\vspace{-1mm}
\includegraphics[width=0.95\textwidth]{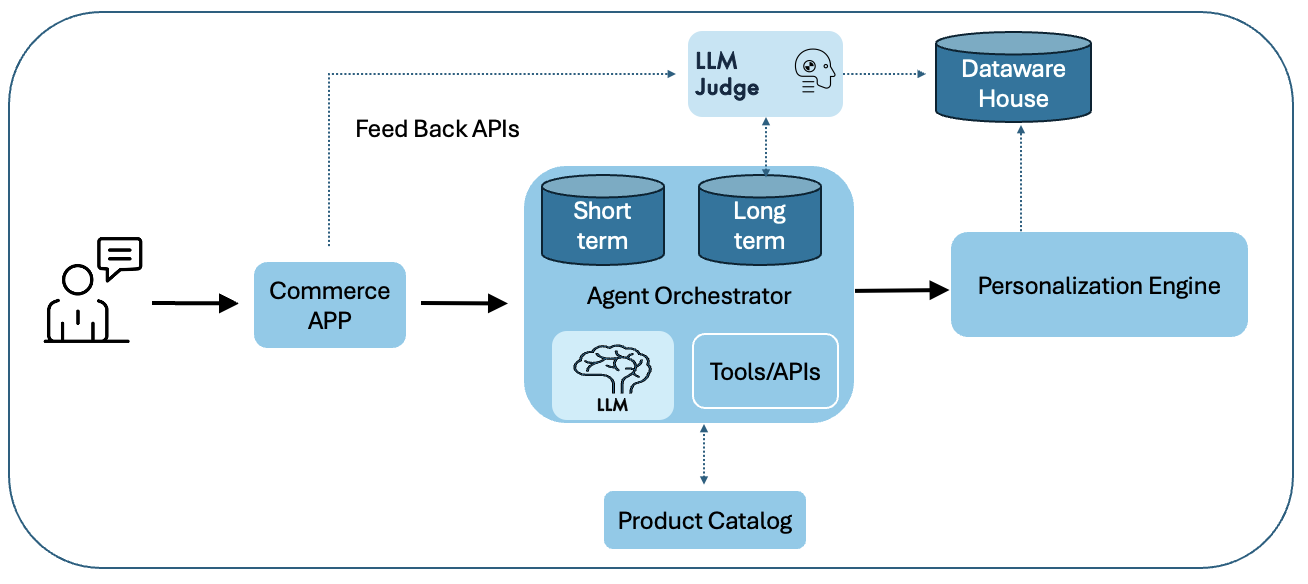}
\vspace{-1.5mm}
\caption{Agentic Commerce AI system: (1) Commerce Platform (2)  Agent Orchestration Framework, (3) LLMs Integration (4) Personalization Engine.}
 
\vspace{-2mm}
\label{fig:agent_system}
\end{figure*}
\subsection{LLM-Powered Commerce Search $\&$ Recommendation}
Our approach utilizes Large Language Models (LLMs), to improve e-commerce search and recommendations through a multi-step process that combines personalized user query understanding with intelligent search method. Our system starts with creating user profiles offline, where a pre-trained LLMs look at user purchase history, demographics, and shopping patterns to build detailed user profiles that capture what each person likes and how they shop. When users search for products, our fine-tuned Nemotron model extracts key attributes and formulates user query , concatenating with  the user's  profile, creating searches  that go beyond simple keyword matching to understand what the user really wants. Our system then uses search engine to find relevant products from catalogs, followed by a ranking step where our another fine-tuned LLMs create personalized Top-K recommendations based on both the found products and the user's preference. This approach changes traditional keyword-based search into a LLM-powered, context-aware system that provides highly personalized shopping experiences by understanding not just what users are looking for, but who they are and what they might prefer based on their shopping behavior. (see Figure \ref{fig:recsystem})

To leverage the powerful text generation and understanding capabilities of large language models (LLMs), we adopt the Hypothetical Document Embeddings (HyDE) \cite{gao2023precise} approach for product retrieval. When a user prompt is given, the LLM first extracts (attribute, value) pairs from the natural language query and reformulates it into a structured format of (product category, attributes, values). The LLM then generates hypothetical product information that captures the user's intent and desired product characteristics. This hypothetical product description, while potentially containing idealized or imprecise details, effectively encodes the relevance patterns of what the user is seeking. The structured query and hypothetical product information are then sent to our search engine, which retrieves matching products from our product catalog by identifying similar real products in the embedding space, effectively grounding the generated hypothetical description to actual inventory while filtering out any inaccurate details through the encoder's dense representation.

We conducted latency experiments on the three stages of our search and recommendation system: (1) Query Understanding, Expansion, and Formulation; (2) Retrieval; and (3) Ranking. Results showed that Query Formulation using our base model created a bottleneck due to its two sequential stages: attribute extraction and query formulation. The entire latency is 4.5s and the Retrieval due to query expansion takes 3.8s. Therefore, in this work, we focus on fine-tuning the query formulation component using NVIDIA Nemotron models (see example \ref{fig:train_ex}). 

\begin{figure}[t]
\centering
\fbox{\begin{minipage}{0.95\columnwidth}
\small
\textcolor{cyan}{\textbf{User Query:}} ``Suggest tech accessories for skiing''
\vspace{2mm}\\
\textcolor{brown}{\textbf{Model Response:}} The system generates four product categories with specific and generic recommendations:
\begin{itemize}
\item \textcolor{brown}{\textbf{Heated Tech Gloves:}} Vertex II, Generic heated gloves
\item \textcolor{brown}{\textbf{Power Banks:}} Veho 10,000mAh, USB-C power bank  
\item \textcolor{brown}{\textbf{Action Cameras:}} GoPro MAX 360, 360 cameras
\item \textcolor{brown}{\textbf{Phone Cases:}} Mount Ready Case, Protective case
\end{itemize}
Each category includes an explanation of relevance to the user's skiing context.
\end{minipage}}
\caption{Example: Query formulation output by LLM and is used for data generation for fine-tuning}
\label{fig:train_ex}
\end{figure}

\subsection{Commerce Agent System Architecture}
To further our agentic commerce initiatives, we developed a comprehensive agentic system to support conversational AI experiences for users through both web and mobile platforms. Our architecture employs a Java-based backend API server, utilizing the LangChain \cite{mavroudis2024langchain} agent orchestration framework to develop intelligent products that leverage multiple language models and coordinate internal knowledge systems, domain services, and existing machine learning solutions. Our system implements a hybrid cloud solution that integrates PayPal's internal technology stack with cloud platform services  to ensure scalability, reliability, and performance for production deployment. (see Figure \ref{fig:agent_system})

In more details, the  CommerceAI Agent Framework employs a sophisticated LLM-based agent mechanism to orchestrate intelligent e-commerce interactions. At its core, the Generic Agent Orchestrator coordinates between multiple specialized components: the LLM Strategy module selects optimal models from the LLM Model Gardens (for instance: NVIDIA NIM Cloud), the Conversation Planning Framework manages multi-turn dialogue flow, and the Agent Task Planner decomposes complex user requests into executable sub-tasks. The framework leverages the Agent Memory Framework to maintain conversation context and user state across sessions, while the Domain Knowledge Center provides access to product catalogs and business logic through standardized protocols (API, MCP, A2A). Agent Tools enable the system to perform concrete actions such as product search, recommendation, and cart management by interfacing with backend services. Throughout execution, the LLM Evaluator continuously monitors agent performance, the Unified Session Management ensures consistency across user interactions, and feedback loops through Feedback APIs and Content Enrichment APIs enable continuous improvement of the agent's capabilities, creating a closed-loop system for adaptive, personalized commerce experiences.

\section{Experimental Results}
\label{sec:exp}

\begin{figure*}[htbp]
\centering
\begin{subfigure}{0.49\textwidth}
    \includegraphics[width=\textwidth]{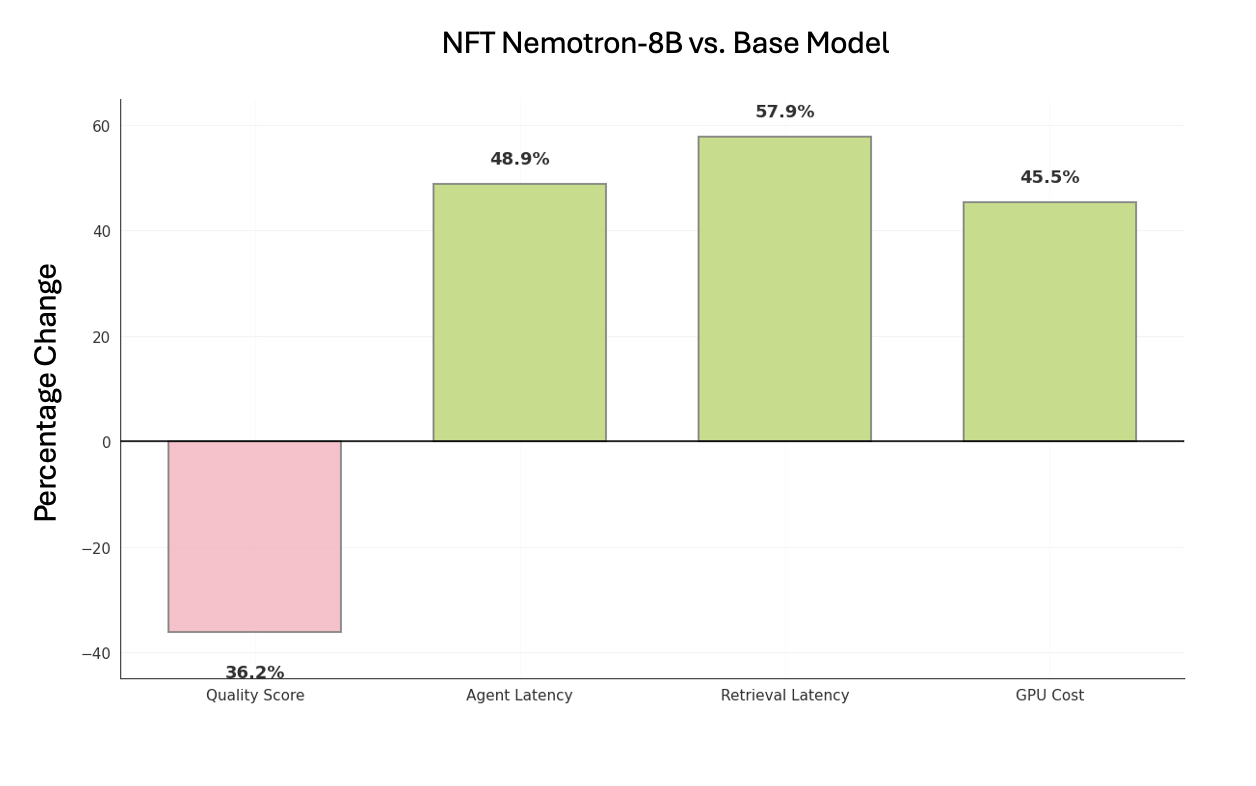}
    \caption{NFT Nemotron-8B shows good performance improvements over base model in latency and cost efficiency, with retrieval latency up 57.9\%, agent latency up 48.9\%, and GPU cost reduced by 45.5\%. However, quality score \footnote{Score is measured on a scale of 0-5 and evaluate the quality of LLM-generated hypothetical products. } decrease 36.2\%, showing a trade-off between speed/cost and output quality.}
    \label{fig:overall2}
\end{subfigure}
\begin{subfigure}{0.49\textwidth}   \includegraphics[width=\textwidth]{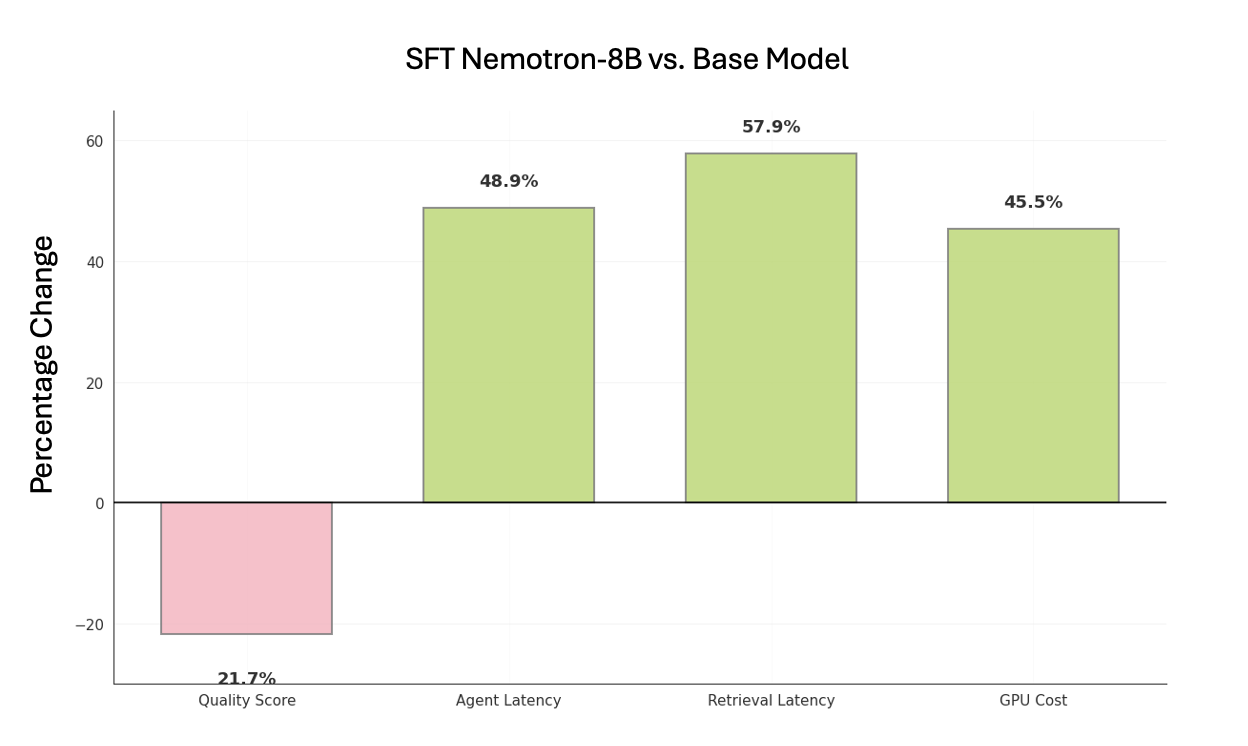}
\caption{SFT Nemotron-8B demonstrates strong performance improvements over our base model, with retrieval latency up 57.9\%, agent latency up 48.9\%, and GPU cost reduced by 45.5\%. The quality score decrease to 21.7\%, showing a better trade-off between efficiency and quality compared to the NFT (a) in above.}
    \label{fig:improvement2}
\end{subfigure}
\caption{Nemotron-8B  Model Performance: SFT vs. NFT over baseline }
\label{fig:all_results_wide}
\end{figure*}

\subsection{Data Sources}
The finetuning dataset is comprised actual customer chat interactions augmented with carefully generated synthetic data to ensure comprehensive coverage of commerce-specific scenarios and edge cases. Specifically, we synthesized 10k prompt–response pairs and produced ground-truth answers using Shopping Assistance Agent data.  This dataset into training:validation $(70\%:30\%)$.   
% A sample of the dataset is added to the Appendix

%  \subsection{Fintune Setup}

% \begin{itemize}
%     \item \textbf{Model:} LLaMA-3.1-Nemotron-Nano-8B-v1
%     \item \textbf{Data Split:} Train = 75\%, Validation = 25\%
%     \item \textbf{Framework:} NVIDIA NeMo SDK
%     \item \textbf{Fine-tuning Method:} Parameter-Efficient Fine-Tuning (PEFT) with LoRA adapters
%     \item \textbf{Hyperparameter Tuning:} Learning rate, LoRA rank/alpha, evaluation frequency
% \end{itemize}

\subsection{Evaluation Methodology}

We finetuned and evaluated multiple model variants with different hyperparameter configurations. Our evaluation employs LLM-as-a-Judge \cite{gu2024survey}, using pairwise and rubric-based comparative evaluations for qualitative analysis, to assess model performance.

\begin{figure}[htbp]
\centering
\begin{subfigure}{0.49\textwidth}
\includegraphics[width=\textwidth]{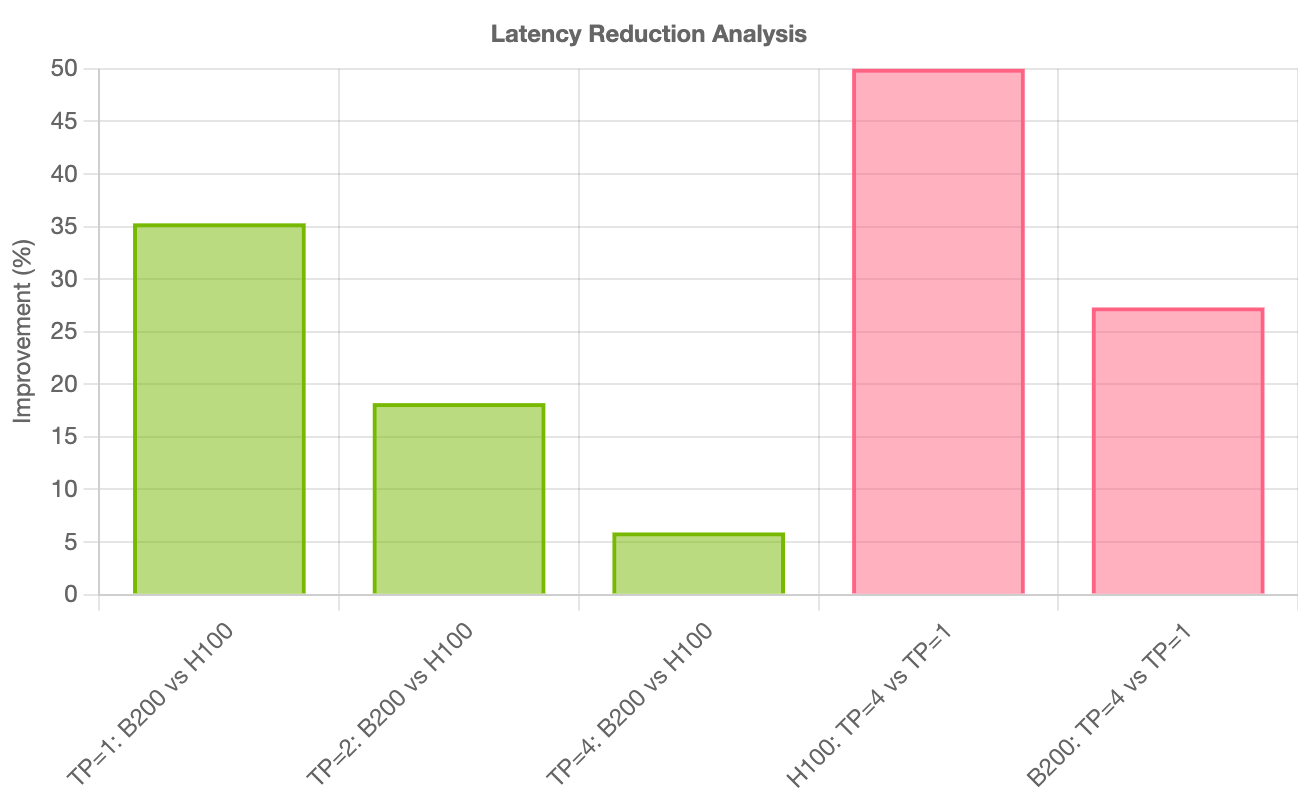}
    % \caption{Improvement Analysis: Percentage improvements in various configurations}
    \label{fig:per14}
\end{subfigure}
\caption{Latency reduction analysis across different configuration. Tensor parallelism configurations (TP=1, TP=2, TP=4) with B200 show  improvements of 35\%, 18\%, and 6\%  compared to H100;  Tensor parallelism (TP=4) to  Tensor parallelism (TP=1) shows latency increases of 50\% for H100 and 27\% for B200}
\label{fig:all_inference_latency}
\end{figure}

\begin{figure*}[htbp]
\centering
\begin{subfigure}{0.79\textwidth}
\includegraphics[width=\textwidth]{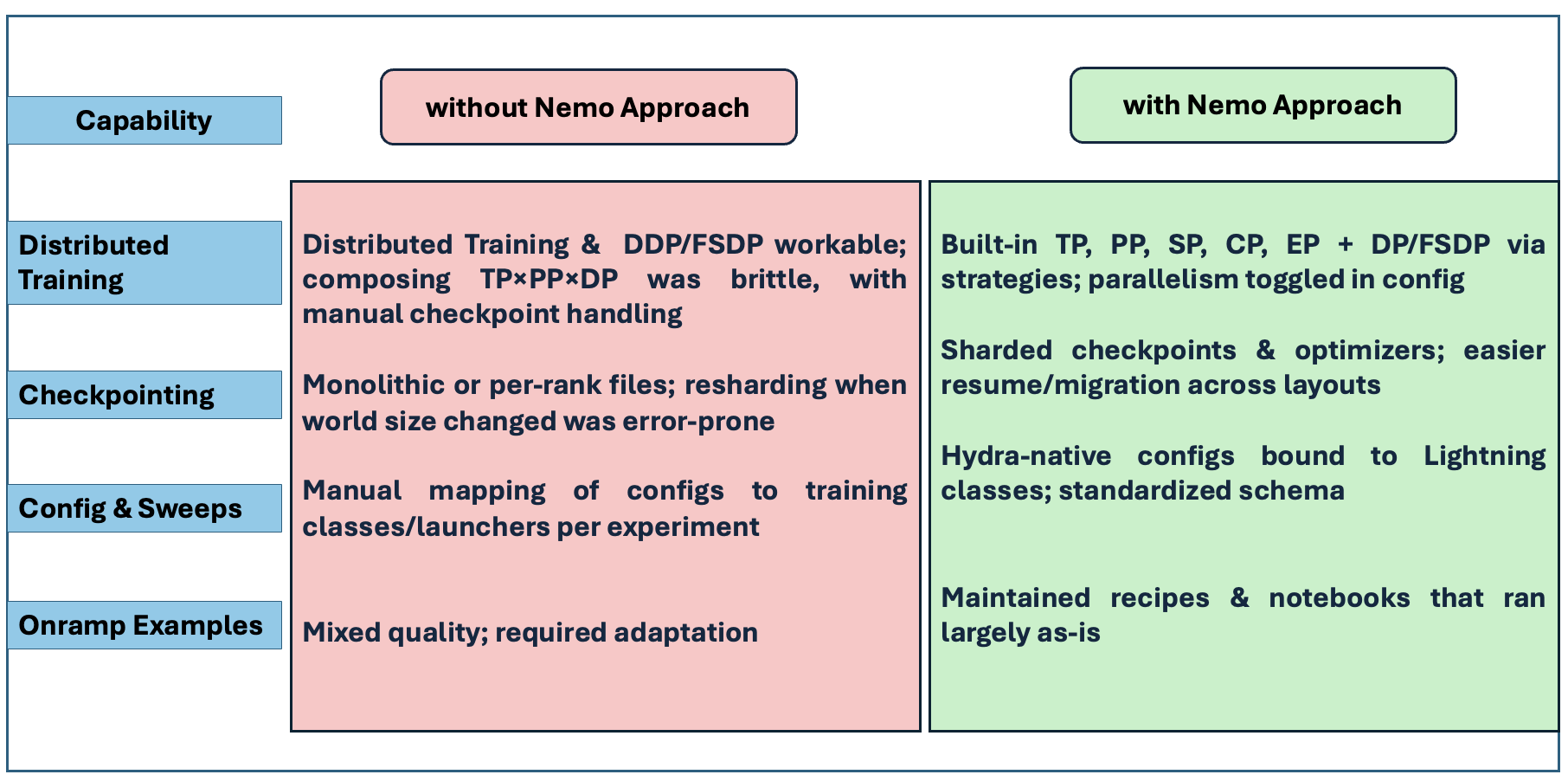}
\end{subfigure}
\caption{ The benefits of Fine-Tuning with NeMo Framework}
\label{fig:comp}
\end{figure*}

\subsection{Supervised Fine Tuning (SFT)}
We leveraged NVIDIA's NeMo SDK as the foundation for fine-tuning a family of small language models (SLMs), focusing on the llama3.1-nemotron-nano-8B-v1 architecture. We conducted comprehensive experiments training 20 LoRA-based model variants while systematically sweeping hyperparameter spaces including learning rates, optimizers (Adam, AdamW), cosine annealing schedules, and LoRA ranks. The NeMo Framework SDK significantly enhanced developer productivity by providing turnkey multi-GPU and multi-dimensional parallelism capabilities, reducing time-to-train through well-maintained training recipes. NeMo's Parameter-Efficient Fine-Tuning (PEFT) support for LoRA/QLoRA techniques \cite{han2024parameter}, combined with sharded checkpoints and distributed optimizers, further simplified experimentation and scale-out operations.

Our experimental evaluation demonstrates the effectiveness of our fine-tuning approach in optimizing the commerce agent's retrieval performance. The SFT \footnote{supervised fine-tuning1} champion model achieved a quality score of 2.49 out of 5 on our challenge dataset, representing a 23\% improvement over the baseline NFT \footnote{without fine-tuning} model (2.03) while maintaining identical inference latency for overall agent response and  retrieval operations. Most significantly, our fine-tuned Nemotron SLM delivers substantial cost and performance benefits compared to our base model, reducing overall agent latency by 49\% , retrieval latency by 58\% , and monthly GPU costs by 45\%  while maintaining competitive quality performance. These results validate our approach of replacing general-purpose LLMs with domain-specific fine-tuned models for specialized commerce tasks, achieving the critical sub-2-second performance target while significantly reducing operational costs. The results are visualized in Figure \ref{fig:all_results_wide}.

\subsection{Direct Preference Optimization (DPO)}

Direct Preference Optimization (DPO) is a training method that learns from relative preferences rather than absolute labels. Instead of teaching the model with traditional ``correct labels,'' DPO presents the model with pairs of outputs for the same prompt: a preferred (better) response and a rejected (worse) response. In our implementation, the preferred response is assigned rank 0, while the rejected response receives rank 1. This approach allows the model to learn nuanced preferences and generate outputs that better align with human expectations. Our DPO training dataset consists of 5,594 training rows and 1,863 validation rows, providing the model with diverse examples of preferred versus rejected responses across various query types.
\begin{figure}[htbp]
\centering
\begin{subfigure}{0.49\textwidth}
    \includegraphics[width=\textwidth]{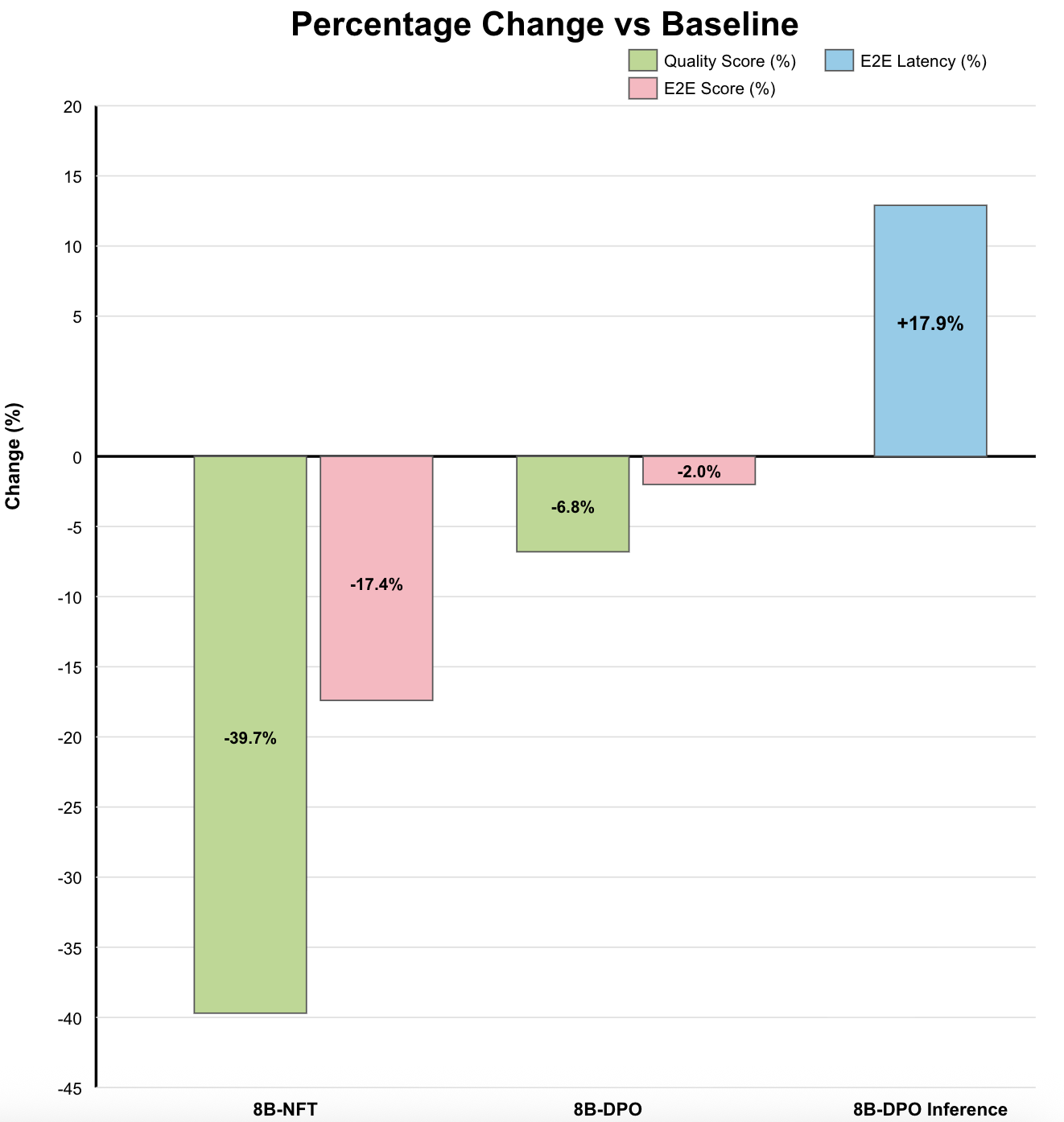}
       \caption{ The 8B-NFT model shows significant degradation in both Quality Score ($-39.7\%$) and E2E Score \footnote{End-to-end score measures overall product recommendation quality across three components: hypothetical product generation, attribute extraction, and product recommendation. } ($-17.4\%$). The 8B-DPO model demonstrates more modest declines in Quality Score ($-6.8\%$) and E2E Score ($-2.0\%$), suggesting improved performance over NFT.  The 8B-DPO Inference configuration achieves a substantial reduction in E2E Latency ($+17.9\%$), indicating faster processing speed. The results highlight the DPO-based approaches minimize quality degradation compared to NFT and inference   delivers meaningful latency improvements at the cost of maintaining accuracy metrics.}
    \label{fig:dpo}
\end{subfigure}
% \begin{subfigure}{0.49\textwidth}   \includegraphics[width=\textwidth]{sections/figures/impact_dpo.png}
% \caption{update!}
%     \label{fig:improvement2}
% \end{subfigure}
% \caption{Nemotron-8B  Model Performance: DPO vs. NFT over baseline }
% \label{fig:all_results_wide}
\end{figure}

\subsection{Inferencing Latency }
NVIDIA NIM is a containerized deployment solution that provides pre-optimized inference microservices with seamless integration to acceleration engines including TensorRT-LLM, vLLM, and SGLang. The platform automatically selects the optimal engine and configuration based on the specific model architecture and underlying hardware characteristics.

In our latency optimization efforts, we explored multiple complementary approaches to achieve sub-2 second performance targets. Our optimization strategy encompassed experimentation with high-performance NVIDIA GPUs including H100 and B200 (Blackwell series), implementation of model partitioning techniques to enable efficient parallelism, and deployment of guided JSON methods for streamlined post-processing operations. 

Our GPU inference benchmarking shows that the NVIDIA B200 Blackwell architecture consistently outperforms the H100 across all tensor parallelism configurations, achieving optimal performance with TP=4. The B200 demonstrates a substantial 35\% latency improvement over H100 at single-GPU deployment (TP=1), while tensor parallelism provides significant performance benefits for both architectures, delivering up to 50\% latency reduction when scaling from TP=1 to TP=4. These results underscore the effectiveness of both advanced GPU architecture improvements and parallelization strategies in optimizing inference performance for our fine-tuned Nemotron model. The results are visualized in Figure \ref{fig:all_inference_latency}

\subsection{SDK Framework Comparison}

Our experience with NVIDIA NeMo SDK demonstrated significant advantages across productivity, scalability, and model fine-tuning workflows: (1) Configuration management: Hydra-first YAML/CLI system enabled parallel tracking of 20 LoRA variants without manual scripting; (2) Scalability: 3D parallelism (TP/PP/DP) with sequence and context parallelism maintained stable, composable training; (3) Checkpoint flexibility: Sharded artifacts simplified cluster migration and parallelism adjustments; (4) PEFT integration: Built-in LoRA/QLoRA support reduced bring-up time and standardized evaluation; (5) Data versatility: Native support for both chat and prompt-response formats accommodated diverse use cases without pipeline restructuring.
(see more in Figure \ref{fig:comp})

% \begin{table*}[htbp]
% \centering
% \caption{NeMo Framework Benefits vs. Alternative Approaches for Fine-Tuning}
% \label{tab:nemo_benefits}
% \small
% \begin{tabular}{|p{2.2cm}|p{5.5cm}|p{5.5cm}|}
% \hline
% \textbf{Capability} & 
% \textbf{Prior Approach (HF/PyTorch Native)} & 
% \textbf{With NeMo} \\
% \hline
% Distributed Training & 
% DDP/FSDP workable; composing TP×PP×DP was brittle, with manual checkpoint handling & 
% Built-in TP, PP, SP, CP, EP + DP/FSDP via strategies; parallelism toggled in config \\
% \hline
% Checkpointing & 
% Monolithic or per-rank files; resharding when world size changed was error-prone & 
% Sharded checkpoints \& optimizers; easier resume/migration across layouts \\
% \hline
% Config \& Sweeps & 
% Manual mapping of configs to training classes/launchers per experiment & 
% Hydra-native configs bound to Lightning classes; standardized schema \\
% \hline
% Onramp Examples & 
% Mixed quality; required adaptation & 
% Maintained recipes \& notebooks that ran largely as-is \\
% \hline
% \end{tabular}
% \label{tab:finding}
% \end{table*}

% \subsection{DPO Alignment}

\section{Conclusion}
This work demonstrates the effectiveness of leveraging NVIDIA's NeMo Framework for optimizing PayPal's Commerce Agent through domain-specific fine-tuning and preference optimization of the llama3.1-nemotron-nano-8B-v1 model. Our systematic experiments with 20 LoRA-based variants achieved significant performance improvements: 49\% reduction in agent latency, 58\% improvement in retrieval latency, and 45\% cost reduction compared to our base model while maintaining competitive quality. We also performed further post-training by incorporating advanced techniques including Direct Preference Optimization (DPO) using Reinforcement Learning (RL). Based on our directional progress, we are confident to establish a comprehensive framework for multi-agent system optimization in production e-commerce environments.

\bibliographystyle{abbrv}
\bibliography{sections/qarsumo}
% \clearpage
% \appendix
% \section{Appendix}
% \input{appendix/sample}

\end{document}